%% file: emnlp2021.tex
\documentclass[11pt,a4paper]{article}
\input{header.tex}

\input{macro.tex}

\title{\contour{black}{\color{golden}GOLD}: Improving Out-of-Scope Detection in \\
Dialogues using Data Augmentation}

\author{Derek Chen \\
  ASAPP, New York, NY 10007 \\
  \texttt{dchen@asapp.com} \\\And
  Zhou Yu \\
  Columbia University, NY \\
  \texttt{zy2461@columbia.edu} \\}

\date{}

\begin{document}
\maketitle
\begin{abstract}
\input{sections/0_abstract.tex}
\end{abstract}

\input{sections/1_intro.tex}
\input{sections/2_related.tex}
\input{sections/3_methods.tex}
\input{sections/4_experiments.tex}
\input{sections/5_results.tex}
\input{sections/6_discussion.tex}

\input{sections/7_conclusion.tex}

\section*{Acknowledgments}
The authors are grateful to Tao Lei, Yi Yang, Jason Wu and Samuel R. Bowman for reviewing earlier versions of the manuscript.  We would also like to thank David Gros and other members of the NLP Dialogue Group at Columbia University for their continued feedback and support.

\bibliography{emnlp2021} 
\bibliographystyle{emnlp_natbib}

\clearpage
\appendix
\input{sections/8_appendix.tex}
\end{document}

%% file: header.tex
\pdfoutput=1

\usepackage{emnlp2021}
\usepackage{times}
\usepackage{latexsym}
\usepackage{import}
\usepackage{xcolor}
\usepackage{contour}
\contournumber{4} 

\usepackage{array}
\usepackage{float}
\usepackage{graphicx}
\usepackage{makecell}
\usepackage{url}
\usepackage{hhline}
\usepackage{hyperref}
\usepackage{amssymb}
\usepackage{amsmath}
\usepackage{array}

\usepackage{arydshln}
\usepackage{algorithm}
\usepackage{algpseudocode}
\makeatletter
\let\OldStatex\Statex
\renewcommand{\Statex}[1][3]{%
  \setlength\@tempdima{\algorithmicindent}%
  \OldStatex\hskip\dimexpr#1\@tempdima\relax}
\makeatother
\algrenewcommand\algorithmicindent{1.3em}

\usepackage{arydshln}
\usepackage{microtype}

\definecolor{golden}{RGB}{255,185,54}


%% file: macro.tex
\newcommand{\D}{\mathcal{D}}  
\newcommand{\T}{\mathcal{T}}  
\newcommand{\A}{\mathcal{A}}  

\newcommand{\ourmethod}{GOLD}

%% file: sections/0_abstract.tex
Practical dialogue systems require robust methods of detecting out-of-scope (OOS) utterances to avoid conversational breakdowns and related failure modes. Directly training a model with labeled OOS examples yields reasonable performance, but obtaining such data is a resource-intensive process. 
To tackle this limited-data problem, previous methods focus on better modeling the distribution of in-scope (INS) examples. 

We introduce \ourmethod~as an orthogonal technique that augments existing data to train better OOS detectors operating in low-data regimes. GOLD generates pseudo-labeled candidates using samples from an auxiliary dataset and keeps only the most beneficial candidates for training through a novel filtering mechanism. In experiments across three target benchmarks, the top GOLD model outperforms all existing methods on all key metrics, 
achieving relative gains of 52.4\%, 48.9\% and 50.3\% against median baseline performance. We also analyze the unique properties of OOS data to identify key factors for optimally applying our proposed method.\footnote{All code and data for major experiments are available at \href{https://github.com/asappresearch/gold}{\texttt{https://github.com/asappresearch/gold}}}

%% file: sections/1_intro.tex
\section{Introduction}

Detecting out-of-scope scenarios is an essential skill of dialogue systems deployed into the real world.  While an ideal system would behave appropriately 
in all conversational settings, such perfection is not possible given that training data is finite, while user inputs are not~\cite{geiger2019posing}.  
Out-of-distribution issues occur when the model encounters situations not covered during training, including novel user intents, domain shifts or custom entities~\cite{kamath2020selective, cavalin2020improving}.  Unique to conversations, dialogue breakdowns represent cases where the user cannot continue the interaction with the system, perhaps due to ambiguous requests or prior misunderstandings~\cite{martinovsky2003breakdown, higashinaka16breakdown}. Such breakdowns might fall within the distribution of plausible utterances, yet still fail to make sense due to the given context.  OOS detection aims to recognize both out-of-distribution problems and dialogue breakdowns.

\begin{figure}
  \includegraphics[width=\linewidth]{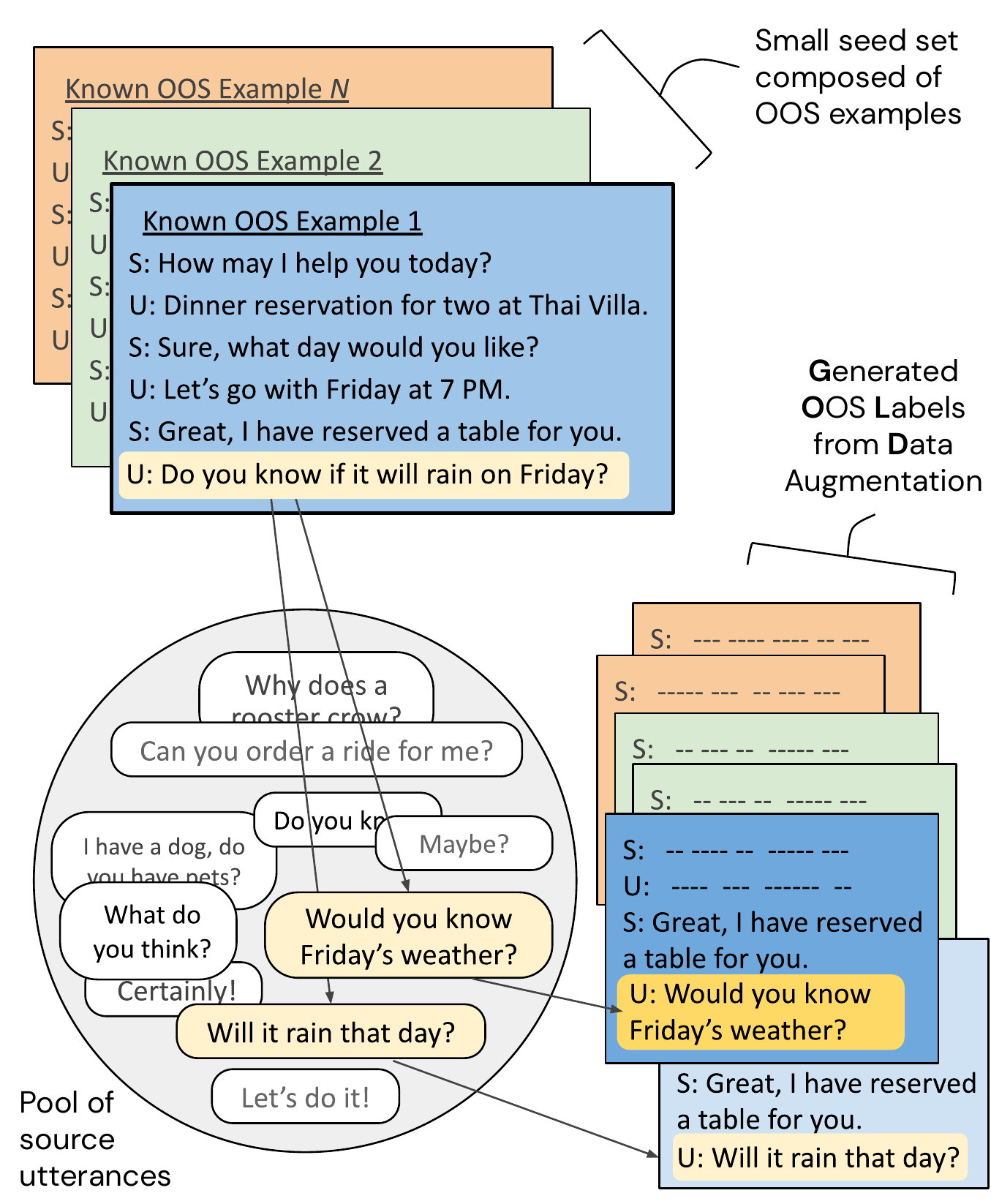}
  \caption{GOLD performs data augmentation by extracting utterances from a source dataset and merging those sentences with known OOS samples from the target dataset to generate pseudo-labeled OOS examples.}
  \label{fig:front_page}
\end{figure}


Prior methods tackling OOS detection in text have shown great promise, but typically assume access to a sufficient amount of labeled OOS data during training~\cite{larson2019evaluation}, which is unrealistic in open-world settings~\cite{fei2016openworld}. Alternative methods
have also been explored which train a supporting model using in-scope data rather than directly training a core model to detect OOS instances~\cite{gangal2020likelihood}.  As a result, they suffer from a mismatch where the objective during training does not line up with the eventual inference task, likely leading to suboptimal performance. 

More recently, data augmentation techniques have been applied to in-scope (INS) data to improve out-of-domain robustness~\cite{ng2020ssmba, zheng2020outofdomain}. 
However, we hypothesize that 
since INS data comes from a different distribution as OOS data, augmentation on the former will not perform as well as augmentation on the latter.

In this paper, we propose a method of \textbf{G}enerating \textbf{O}ut-of-scope \textbf{L}abels with \textbf{D}ata augmentation (\ourmethod) to improve OOS detection in dialogue. To create new pseudo-labeled examples, we start with a small seed set of known OOS examples.  Next, we find utterances that are similar to the known OOS examples within an auxiliary dataset.  We then generate candidate labels by replacing text from the known OOS examples with the similar utterances uncovered in the previous step. Lastly, we run an election to filter down the candidates to only those which are most likely to be out-of-scope. Our method is complementary to other indirect prediction techniques and in fact takes advantage of progress by other methods. 

We demonstrate the effectiveness of \ourmethod~across three task-oriented dialogue datasets, where our method achieves state-of-the-art performance across all key metrics.  We conduct extensive ablations and additional experiments to probe the robustness of our best performing model.  Finally, we provide analysis and insights on augmenting OOS data for other dialogue systems.

%% file: sections/2_related.tex
\section{Related Work}

\subsection{Direct Prediction}
A straightforward method of detecting out-of-scope scenarios is to train directly on OOS examples~\cite{fumera2003classification}.  These situations are encountered more broadly by the insertion of any out-of-distribution response or more specifically 
when a particular utterance does not make sense in the current context. 

\paragraph{Out-of-Distribution Recognition} An utterance may be out-of-scope because it was not included in 
the distribution the dialogue model was trained on.  Distribution shifts may occur due to unknown user intents, different domains or incoherent speech.
We differ from such methods since they either operate on images~\cite{kim2018joint, hendrycks2019outlier, mohseni2020selfsupervised} 
or assume access to an impractically large number of OOS examples in relation to INS examples~\cite{tan2019outofdomain, kamath2020selective, larson2019evaluation}  

\paragraph{Dialogue Breakdown}
In comparison to out-of-distribution cases, dialogue breakdowns are unique to conversations because they depend on context~\cite{higashinaka16breakdown}. In other words, the utterances fall within the distribution of reasonable responses but are out-of-scope due to the state of the particular dialogue. Such breakdowns occur when the conversation can no longer proceed smoothly due to an ambiguous statement from the user or some misunderstanding made by the agent~\cite{ng2020breakdown}.  \ourmethod~also focuses on dialogue, but additionally operates under the setting of limited access to OOS data during training~\cite{hendriksen2019dialogue}.  

\subsection{Indirect Prediction}
\label{subsection:related}
An alternative set of methods for OOS detection assume access to a supporting model trained solely on in-scope data.  There are roughly three ways in which a core detector model can take advantage of the pre-trained supporting model.

\paragraph{Probability Threshold} The first class of methods utilize the output probability of the supporting model to determine whether an input is out-of-scope.  More specifically, if the supporting model's maximum output probability falls below some threshold $\tau$, then it is deemed uncertain and the core detector model labels the input as OOS~\cite{hendrycks2017baseline}.  
The confidence score of the supporting model can also be manipulated in a number of ways to help further separate the INS and OOS examples~\cite{liang2018odin, lee2018training}. 
Other variations include setting thresholds on reconstruction loss~\cite{ryu2017neural}  or on likelihood ratios ~\cite{ren2019likelihood}. 

\paragraph{Outlier Distance} Another class of methods define out-of-scope examples as outliers whose distance is far away from known in-scope examples~\cite{gu2019nearest, mandelbaum2017distance}. Variants can tweak the embedding function or distance function used for determining the degree of separation.~\cite{cavalin2020improving, oh2018outofdomain, yilmaz2020kloos}. For example, Local Outlier Factor (LOF) defines an outlier as a point whose density is lower than that of its nearest neighbors~\cite{breunig2000lof, lin2019deep}.

\paragraph{Bayesian Ensembles} The final class of methods utilize the variance of supporting models to make decisions.
When the variance of the predictions is high, then the input is supposedly difficult to recognize and thus out-of-distribution.  Such ensembles can be formed explicitly through a collection of models~\cite{vyas2018outofdistribution, shu2017doc, lakshminarayanan2017simple} or implicitly through multiple applications of dropout~\cite{gal2016dropout}.

\subsection{Data Augmentation}
Our method also pertains to the use of data augmentation to improve model performance under low resource settings.

\paragraph{Augmentation in NLP}
Data augmentation for NLP has been studied extensively in the past~\cite{jia2016recombination, silfverberg2017data, furstenau2009srl}. Common methods include those that alter the surface form text~\cite{wei2019eda} or perturb a latent embedding space ~\cite{wang2015annoying, fadaee2017nmt, liu2020databoost}, as well as those that perform paraphrasing~\cite{zhang19paws}.  
Alternatively, masked language models generate new examples by proposing context-aware replacements for the masked token~\cite{kobayashi2018contextual, wu2019conditional}.

\paragraph{Data Augmentation for Dialogue} Methods for augmenting data to train dialogue systems are most closely related to our work.
Previous research has used data augmentation to improve natural language understanding (NLU) and intent detection in dialogue~\cite{niu2019automatically, hou2018sequence}. 
Other methods augment the in-scope sample representations to support out-of-scope robustness~\cite{ryu2018outofdomain, ng2020ssmba, lee2019contextual}. Recently, generative adversarial networks (GANs) have been used to create out-of-domain examples that mimic known in-scope examples~\cite{zheng2020outofdomain, marek2021oodgan}.  In contrast, we operate directly on OOS samples and consciously generate data far away from anything seen during pre-training, a decision which our later analysis reveals to be quite important.

%% file: sections/3_methods.tex
\section{Background and Baselines}
In this section we formally describe the task of out-of-scope detection and the different approaches to handling this issue.

\subsection{Problem Formulation}
Let $\mathcal{D}_{direct} = \{ (x_1, y_1), ... ,(x_n, y_n) \}$ be a target dataset containing a mixture of in-scope and out-of-scope dialogues.  The input context $x_i = \{(S_1, U_1), ..., (S_t, U_t)\}$ is a series of system and user utterances within $t$ turns of a conversation.  The desired output $y_i \in [0,1]$ is a binary label representing whether that context is out-of-scope. We define OOS to encompass both out-of-distribution utterances, such as out-of-domain intents or gibberish speech, as well as in-distribution utterances spoken in an ambiguous manner. A model given access to such a dataset is an OOS detector $P_{\theta}(y_i|x_i)$ performing \textit{direct} prediction.

In contrast, the problem we tackle in this paper is \textit{indirect} prediction, where only a limited or nonexistent number of OOS examples are available during training. Instead, the training data is sampled 
from in-scope dialogues $\mathcal{D}_{indirect} \sim \mathcal{P}_{INS}$, and the labels $y_j \in \mathcal{Y}$ represent a set of known user intents.  This data may be used to train an intent classifier which then acts as a supporting model to the core OOS detector during inference.  Critically, the supporting model $P_{\psi}(y_j|x_i)$ has never encountered out-of-scope utterances during training.

\begin{figure*}[t]
\includegraphics[width=\textwidth]{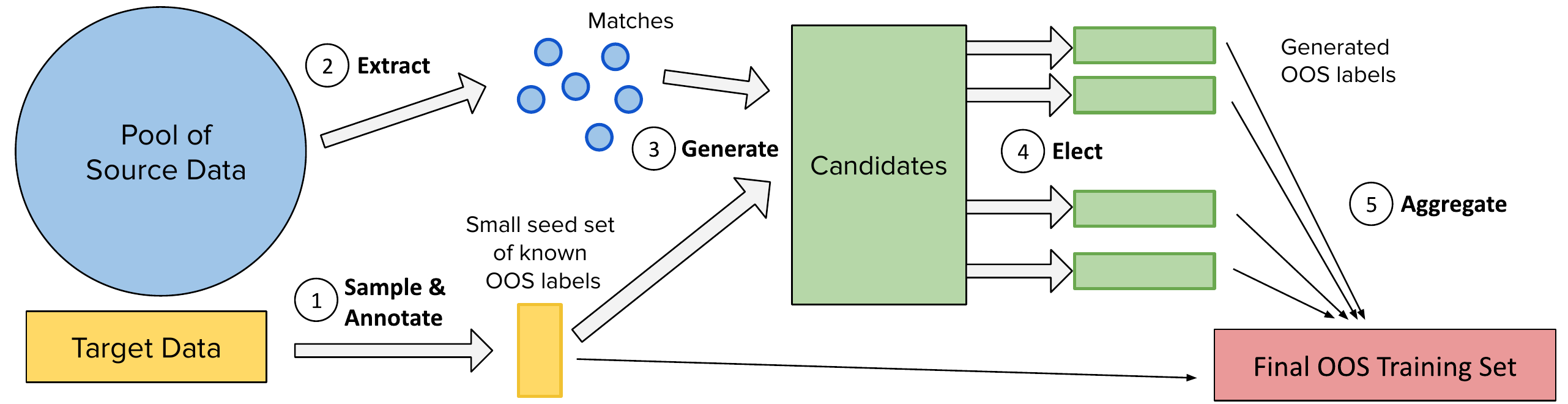}
\caption{Full GOLD pipeline: (1) Sample and annotate a small seed set from unlabeled target data.
(2) Extract similar matches from the source dataset. 
(3) Generate candidates by swapping utterances in the seed data with match sentences.
(4) Elect the top candidates to become pseudo-labeled OOS examples.
(5) Aggregate all elected labels to form the final OOS training set.}
\centering
\label{fig:pipeline}
\end{figure*}

\subsection{Baselines}
Prior methods for approaching indirect prediction generally fall into three categories: probability threshold, outlier distance and Bayesian ensemble.  In all cases, the supporting model trained on the intent classification task uses a pretrained BERT model as its base~\cite{devlin2019bert}.

Starting with \textit{Probability Threshold} baselines, (1) \textbf{MaxProb} declares an example as OOS if the maximum value of the supporting model's output probability distribution falls below some threshold $\tau$~\cite{hendrycks2017baseline}. (2) \textbf{ODIN} enhances this by adding temperature scaling and small perturbations to the input which help to increase the gap between INS and OOS instances~\cite{liang2018odin}. (3) \textbf{Entropy} considers an example to be OOS if the supporting model is uncertain, as determined by the entropy level rising above a threshold $\tau$~\cite{lewis1994sequential}.

\textit{Outlier Distance} baselines find OOS examples by casting the problem as detecting outliers.  Inputs are considered outliers when their embeddings are too far away from clusters of INS embeddings as measured by some threshold $\tau$.
The (4) \textbf{BERT} baseline embeds utterances uses the supporting model pre-trained on intent classification and measures separation by Euclidean distance. Based on the success in ~\cite{podolskiy2021revisiting}, the (5) \textbf{Mahalanobis} method embeds examples with a vanilla RoBERTa model and uses the Mahalanobis distance~\cite{liu2019roberta}.  Finally, inspired by BADGE for active learning~\cite{ash2020badge}, the (6) \textbf{Gradient} method sets the embedding of each example as the gradient vector of the input tokens as computed by back-propagation.  

\textit{Bayesian Ensembles} predict labels by the amount of variation formed by the estimates of an ensemble. More specifically, (7) \textbf{Dropout} implicitly creates a new model whenever it randomly drops a percentage of its nodes~\cite{gal2016dropout}.  
During inference, each input is passed through the supporting model $k$ times to estimate the user intent.  If the ensemble fails to reach a majority vote on the intent classification task, then the example is assigned as out-of-scope. 

\section{\ourmethod}
To avoid a mismatch between training and inference, we are motivated to explore the direct prediction paradigm in a way the does not violate the OOS data restriction inherent to indirect prediction methods.
Concretely, \ourmethod~performs data augmentation on a small sample of labeled OOS examples to generate pseudo-OOS data.  This weakly-labeled data is then combined with INS data for training a core OOS detector.  We limit the number of OOS samples to be only 1\% of the size of in-scope training examples.  Note that indirect methods also typically have access to a modest number of OOS samples for tuning hyper-parameters, such as thresholds, so this adjustment is not an exclusive advantage of our method. 

In addition to a small seed set of OOS examples, we assume access to an external pool of utterances, which serve as the source of data augmentations, similar to \citet{hendrycks2019outlier}.  We refer to this auxiliary data as the \textit{source} dataset $\mathcal{S}$, as opposed to the \textit{target} dataset $\mathcal{T}$ used for evaluating our method.  \ourmethod~now proceeds in three basic steps. (See Algorithm \ref{alg:method} for full details.)


\subsection{Match Extraction} Our first step is to find utterances in the source data that closely match the examples in the OOS seed data.  We encode all source and seed data into a shared embedding space to allow for comparison.  When the seed example is a multi-turn dialogue, we embed only the final user utterance.  Then for each seed utterance, we extract $d$ similar utterances from source $\mathcal{S}$ as measured by cosine distance,\footnote{We also considered Euclidean distance and found that to yield negligible difference in preliminary testing.} where $d$ is the desired number of matches.  For example, 
as seen in Figure~\ref{fig:front_page}, the seed text ``\textit{Do you know if it will rain on Friday?}'' extracts ``\textit{Will it rain that day?}'' as a match.  We discuss different types of embedding mechanisms in section \ref{subsection:experiments}.

\subsection{Candidate Generation} Since dialogue contexts often contain multiple utterances, we want our augmented examples to also span multiple turns. Accordingly, our next step involves generating candidates by carefully crafting new conversations using the existing dialogue contexts in the seed data.  Each new candidate is formed by swapping a random user utterance in the seed data with a match utterance from the source data. Notably, agent utterances in the seed data are left untouched during this process. 

\subsection{Target Election} Candidates are merely pseudo-labeled as OOS, so relying on such data as a training signal might be quite noisy.  Accordingly, we apply a filtering mechanism to ensure that only the candidates most likely to be out-of-scope are ``elected'' to become target OOS data.  Elections are held by running all the candidates through an ensemble of baseline detectors.  Specifically, we choose the top detectors from each of the major indirect prediction categories which results in three voters.  If the majority of voters agree that an example is out-of-scope, then we include that candidate in our target pool.  

As a last step, we aggregate the pseudo-labeled OOS examples, the small seed set of known OOS examples and the original INS examples to form the final training set for our model. We train a classifier with this data to directly predict out-of-scope instances.


    
\begin{algorithm}[H] 
\begin{algorithmic}[1]
    \Require \parbox[t]{\dimexpr\linewidth-\algorithmicindent}
            {ensemble of baseline detectors $e$ \\
            external source dataset $\mathcal{S}$ \strut}
    \State \textbf{Input:} \parbox[t]{\dimexpr\linewidth-\algorithmicindent}
            {Labeled, in-scope data from target data \\
            $\T = \{(x_1, y_1)\ldots(x_n, y_n)\}$ \\
            Unlabeled data from target distribution \\
            $\T' = \{(x_1)\ldots(x_m)\}$ \\
            Desired number of matches $d$ \strut}
    \Function{SwapAugment}{$\T, \T', d$}
    \State seed set $\A \leftarrow$ sample and annotate $\T'$
    \State $\mathcal{S}' \leftarrow$ embed all items in $\mathcal{S}$
    \For {instance $i \in \A$}:
        \State initialize $A_i = \{ \}$
        \While{size$(A_i) < d$:}
            \State $i' \leftarrow$ embed instance $i$
            \State extract $m$ nearest neighbors of $i'$
            \Statex[3] from $\mathcal{S}'$ by cosine distance
            \For {$j \in m$ matches}:
                \State candidate $c \leftarrow$ generate($j$, $i$)
                \State votes $\leftarrow$ ensemble $e$ holds an
                \Statex[4] election on candidate $c$
                \If {majority(votes)}:
                    \State {$A_i \leftarrow A_i \cup c$}
                \EndIf
            \EndFor
        \EndWhile
    \EndFor
    \State $A'$ = aggregate($A_i$)
    \State augmented dataset $\D \leftarrow \T \cup \A \cup \A'$
    \State \Return $\D$ 
    \EndFunction
\end{algorithmic}
\caption{\ourmethod}
\label{alg:method}
\end{algorithm}

%% file: sections/4_experiments.tex
\section{Experimental Setup}

\subsection{Target Datasets}
\label{subsection:target}
We test our detection method on three dialogue datasets. Example counts shown in Table~\ref{tab:targets}.

\paragraph{\textit{\underline{S}}chema-guided Dialog Dataset for \textit{\underline{T}}ransfer Le\textit{\underline{ar}}ning} STAR is a task-oriented dataset containing 6,651 multi-domain dialogues with turn-level intents~\cite{mosig2020star}.  Following the suggestion in Section 6.3 of their paper, we adapt the data for out-of-domain detection by selecting responses labeled as ``ambiguous'' or ``out-of-scope'' to serve as OOS examples.  After filtering out generic utterances (such as greetings), we are left with 29,104 examples consisting of 152 user intents.  Since the corpus does not strictly define a train and test set, we perform a random 80/10/10 split of the dialogues and other minor pre-processing to prepare the data for training.

\begin{table}
\centering
\resizebox{\linewidth}{!}{
\begin{tabular}{l|ccc}
\Xhline{1pt}
\textbf{Split} & \textit{\textbf{STAR}} & \textit{\textbf{FLOW}} & \textit{\textbf{ROSTD}} \\
\hline
Train   & 22,051/1,248    & 60,119/4,499    & 30,521/3,200  \\
Dev     & 2,751/178      & 3,239/228      & 4,181/453    \\
Test    & 2,708/168      & 3,227/239      & 8,621/937    \\
\Xhline{1pt}
\end{tabular}}
\caption{Data count for each target dataset, broken down by number of in-scope/out-of-scope examples.} \label{tab:targets}
\end{table}

\paragraph{SM Calendar \textit{\underline{Flow}}} FLOW is also a task-oriented dataset with turn-level annotations~\cite{andreas-etal-2020-task}. Originally built for semantic parsing, FLOW is structured as a novel dataflow object that takes
the form of a computational graph.  For our purposes, we take advantage of the `Fence' related labels found in the dataset, which represent situations where a user is straying too far away from discussions within the scope of the system, and thus need to be ``fenced-in''.
We focus on utterances associated with a clear intent, once again dropping turns representing greetings and other pleasantries, which results in 71,551 examples spanning 44 total intents.  The test set is hidden behind a leaderboard, so we divide the development set in half, resulting in an approximate 90/5/5 split for train, dev and test, respectively.

\paragraph{\textit{\underline{R}}eal \textit{\underline{O}}ut-of-Domain \textit{\underline{S}}entences From \textit{\underline{T}}ask-oriented \textit{\underline{D}}ialog} ROSTD is a dataset explicitly designed for out-of-distribution recognition~\cite{gangal2020likelihood}.  The authors constructed sentences to be OOS examples with respect to a separate dataset collected by~\citet{schuster2019cross}.  The dialogues found in the original dataset then represent the INS examples.  ROSTD contains 47,913 total utterances spanning 13 intent classes and comes with a pre-defined 70/10/20 split which we leave unaltered. The dataset is less conversational since each example consists of a single turn command, while its labels are higher precision since each OOS instance is human-curated.

\subsection{Evaluation Metrics}
\label{subsection:metrics}
Following prior work on out-of-distribution detection~\cite{hendrycks2017baseline, ren2019likelihood}, we evaluate our method on three primary metrics. (1) Area under the receiver operating characteristic curve (AUROC) measures the probability that a random OOS example will have a higher probability of being out-of-scope than 
a randomly selected INS example~\cite{davis2016auroc}.  This metric averages across all thresholds and is therefore threshold independent. (2) The area under the precision-recall curve (AUPR) is another holistic metric which summarizes performance across multiple thresholds.  The AUPR is most useful in scenarios containing class imbalance~\cite{manning2001foundations}, which is precisely our case since INS examples greatly outnumber OOS examples. 
(3) The false positive rate at recall of N (FPR@N) is the probability that an INS example raises a false alarm when $N\%$ of OOS examples are detected~\cite{hendrycks2019outlier}.  Thus, unlike the first two metrics, a lower FPR@N is better.  We report FPR at values of N=\{0.90, 0.95\}.

\subsection{Experiments on Model Variants}
\label{subsection:experiments}
In addition to testing against baseline methods, we also run experiments to study the impact of varying the auxiliary dataset and the extraction options.

\subsubsection{Source Datasets}
We consider a range of datasets as sources of augmentation, starting with known out-of-scope queries (OSQ) from the Clinc150 dataset~\cite{larson2019evaluation}. Because our work falls under the dialogue setting, we also consider Taskmaster-2 (TM) as a source of task-oriented utterances~\cite{byrne2019taskmaster} and PersonaChat (PC) for examples of informal chit-chat~\cite{zhang2018personachat}.  Upon examining the validation data, we note that many examples of OOS are driven by users attempting to ask questions that the agent is not able to handle. Thus, we also include a dataset composed of questions extracted from Quora (QQP)~\cite{iyer2017qqp}. Finally, we consider mixing all four datasets together into a single collection (MIX).

\subsubsection{Extraction Techniques}
To optimize the procedure of extracting matches from the source data, we try four different mechanisms for embedding utterances. (1) We feed each OOS instance into a SentenceRoBERTa model pretrained for \textbf{paraphrase} retrieval to find similar utterances within the source data~\cite{reimers2019sentencebert}. (2) As a second option, we encode source data using a static BERT \textbf{Transformer} model~\cite{devlin2019bert}.  Then for each OOS example encoded in the same manner, we extract the nearest source utterances. (3) We embed OOS and source data as a bag-of-words where each token is a 300-dim \textbf{GloVe} embedding~\cite{pennington2014glove}. (4) As a final variation, we embed all utterances with \textbf{TF-IDF} embeddings of 7000 dimensions.  The spectrum of extraction techniques aim to progress from methods that capture strong semantic connections to the OOS seed data towards options with weaker relation to original seed data.

%% file: sections/5_results.tex
\section{Key Results}

\begin{figure}
  \includegraphics[width=\linewidth]{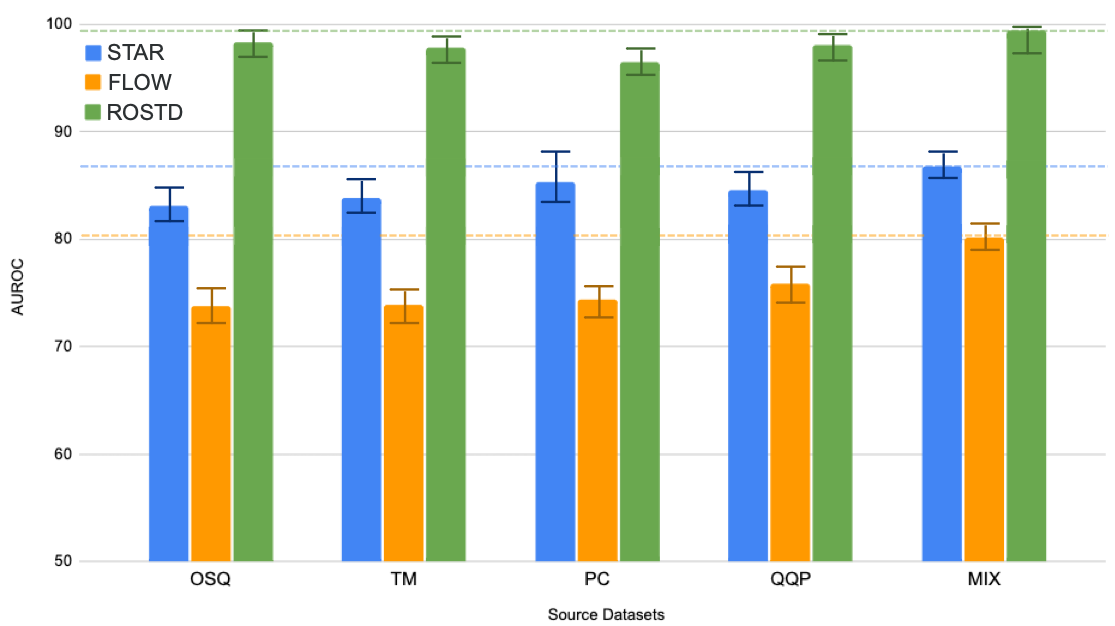}
  \caption{AUROC performance across source datasets}
  \label{fig:source}
\end{figure}

We now present the results of our main experiments. As evidenced by Figure~\ref{fig:source}, MIX performed as the best data source across all datasets, so we use it to report our main metrics within Table \ref{tab:joint}. Also, given the strong performance of GloVe extraction technique across all datasets, we select this version for comparison purposes in the following analyses.

\subsection{STAR Results}
Left columns of Table \ref{tab:joint} present STAR results. Models trained with augmented data from \ourmethod~consistently outperform all other baselines across all metrics.   The top model exhibits gains of 8.5\% in AUROC and 40.0\% in AUPR over the nearest baseline.  Performance is even more impressive in lowering the false positive rate with improvements of 24.2\% and 29.8\% at recalls of 0.95 and 0.90, respectively. Among the different baselines, we observe the Outlier Distance methods generally outperforming the others, with the Mahalanobis method doing the best.  Among \ourmethod~variations, there are mixed results as GloVe and TF-IDF both produce high overall accuracy.  Notably, the Paraphrase method meant to extract matches most similar to the seed data performed the worst.

\begin{table*}
\centering
\resizebox{\textwidth}{!}{
\begin{tabular}{l|cc>{\centering}p{3.5em}>{\centering}p{3.5em}|
cc>{\centering}p{3.5em}>{\centering}p{3.5em}|cc>{\centering}p{3.3em}c}
\Xhline{1pt} & \multicolumn{4}{c}{\textbf{STAR Data}} &
\multicolumn{4}{c}{\textbf{FLOW Data}} & \multicolumn{4}{c}{\textbf{ROSTD Data}} \\
\textbf{Methods} &
\textit{AUROC} & \textit{AUPR} & \textit{FPR@.95} & \textit{FPR@.90} &
\textit{AUROC} & \textit{AUPR} & \textit{FPR@.95} & \textit{FPR@.90} &
\textit{AUROC} & \textit{AUPR} & \textit{FPR@.95} & \textit{FPR@.90} \\

\hline
Oracle      &  0.9869 & 0.8837 & 11.2\% & 2.6\%  
            &  0.8931 & 0.6471 & 62.5\% & 39.8\%   
            &  0.9999 & 0.9992 & 0.03\% & 0.02\%   \\
\hline
MaxProb     &  0.6891 & 0.1824 & 85.8\% & 72.9\%  
            &  0.6881 & 0.1581 & 75.4\% & 67.8\%   
            &  0.7969 & 0.4554 & 54.3\% & 54.2\%   \\
ODIN        &  0.7012 & 0.1860 & 87.3\% & 75.3\%  
            &  0.6893 & 0.1714 & 76.9\% & 69.5\%
            &  0.8087 & 0.4938 & 54.3\% & 54.2\%  \\
Entropy     &  0.7206 & 0.1915 & 87.0\% & 75.7\%
            &  0.6875 & 0.1887 & 77.4\% & 70.2\%
            &  0.8125 & 0.5250 & 54.3\% & 54.2\%  \\
\hdashline
BERT        &  0.7170 & 0.1958 & 85.3\% & 74.1\%
            &  0.5570 & 0.1685 & 98.6\% & 96.0\%
            &  0.9754 & 0.8126 & 8.80\% & 4.45\%   \\ 
Mahalanobis &  0.8002 & 0.3179 & 73.9\% & 58.2\%
            &  0.7004 & 0.1757 & 82.7\% & 72.1\%
            &  0.9583 & 0.7338 & 19.1\% & 11.7\%   \\             
Gradient    &  0.7255 & 0.1402 & 72.0\% & 62.7\%
            &  0.7223 & 0.1850 & 81.4\% & 70.0\%
            &  0.9821 & 0.8729 & 7.97\% & 3.87\%   \\
\hdashline
Dropout     &  0.5332 & 0.0631 & 99.9\% & 99.8\%
            &  0.5091 & 0.0707 & 99.9\% & 99.8\%
            &  0.5036 & 0.1991 & 99.9\% & 99.8\%  \\
\hline
Paraphrase  &  0.8537 & 0.4133 & \underline{62.6\%} & 47.5\% 
            &  0.7767 & 0.2743 &  72.8\% & 60.4\%
            &  0.9967 & 0.9897 &  0.26\%  & 0.16\%  \\
Transformer &  0.8542 & 0.4251 & 65.9\%  & 45.7\%
            & \textbf{0.8059}  & \underline{0.3228}  & \underline{62.7\%} & \underline{51.3\%}
            &  0.9981 & \underline{0.9904} & \underline{0.21\%}  & \textbf{0.09\%}  \\
GloVe       & \textbf{0.8683}  & \underline{0.4450}  & \textbf{56.0\%}           & \underline{40.9\%}
            & \underline{0.8022} &  \textbf{0.3243}  & \textbf{60.6\%}           & \textbf{49.5\%}
            &  \textbf{0.9990} & \textbf{0.9933}     & \textbf{0.17\%} & \textbf{0.09\%} \\
TF-IDF      & \underline{0.8614} &  \textbf{0.4539}  & 68.1\%  & \textbf{40.3\%}
            & 0.7790  & 0.2758 & 74.2\%  & 58.0\%
            & \underline{0.9987} & 0.9905 &  0.56\%  &  0.19\% \\
\hdashline
Random      & 0.8531  & 0.4378 & 68.8\%  & 45.4\%
            & 0.7692  & 0.2889 & 73.1\%  & 62.0\%
            & 0.9984  & 0.9893 & 0.40\%  &  0.19\% \\
\Xhline{1pt}
\end{tabular}}
\caption{Experimental results across all target datasets where bold items indicate best results, and underlined items indicate the runner-up. First seven rows are baselines, while the bottom five rows are models trained with GOLD.} \label{tab:joint}
\end{table*}

\subsection{FLOW Results}
Central columns of Table \ref{tab:joint} present results on FLOW data.  Once again, \ourmethod~models outperform all baselines across all metrics.  This time around, there is not an obvious winner among baselines. On the other hand, GloVe stands out as the clear overall top performer, with Transformer following closely behind.  Models trained on data augmented by GloVe show improvements of 11.1\% in AUROC, 71.9\% in AUPR and 19.5\% for FPR@0.95 over the nearest baseline.  We again notice that the Paraphrase variation does not perform quite as well among \ourmethod~methods.

\begin{table}
\centering
\resizebox{\linewidth}{!}{
\begin{tabular}{l|c>{\centering}p{2.9em}c}
\Xhline{1pt}
\textbf{Methods} & 
\textbf{AUROC}$\uparrow$ & \textbf{AUPR}$\uparrow$ & \textbf{FPR@.95}$\downarrow$ \\
\hline
Likelihood \cite{gangal2020likelihood} &  0.9822 & 0.9647 & 7.41\% \\
OodGAN      \cite{marek2021oodgan}     &  0.9899 & 0.9626 & 2.59\% \\
\hdashline
GOLD w/ GloVe extraction    &  \textbf{0.9990} & \textbf{0.9933} & \textbf{0.17\%} \\  
\Xhline{1pt}
\end{tabular}}
\caption{ROSTD results against previous works} \label{tab:rostd}
\end{table}

\subsection{ROSTD Results}
\label{subsection:results}
As seen in Tables \ref{tab:joint} and \ref{tab:rostd}, \ourmethod~outperforms not only all baselines, but also prior work on ROSTD across all metrics.  
The GloVe method cements its standing at the top with gains of 1.7\% in AUROC, 13.8\% in AUPR and 97.9\% in FPR@0.95 against the top baselines. 
Given the consistently poor performance of Paraphrase yet again,
we conclude that unlike traditional INS data augmentation, augmenting OOS data should \textit{not} aim to find the most similar examples to seed data.  We hypothesize that producing pseudo-labeled OOS data that are too similar to given known-OOS data causes the model to overfit since it is simply optimizing towards the same examples over and over again.


%% file: sections/6_discussion.tex
\section{Discussion and Analysis}
In this section, we conduct follow-up experiments to analyze the impact of our method's components and identify best practices when applying data augmentation for OOS detection.

\subsection{Ablations}

\begin{figure}
  \includegraphics[width=\linewidth]{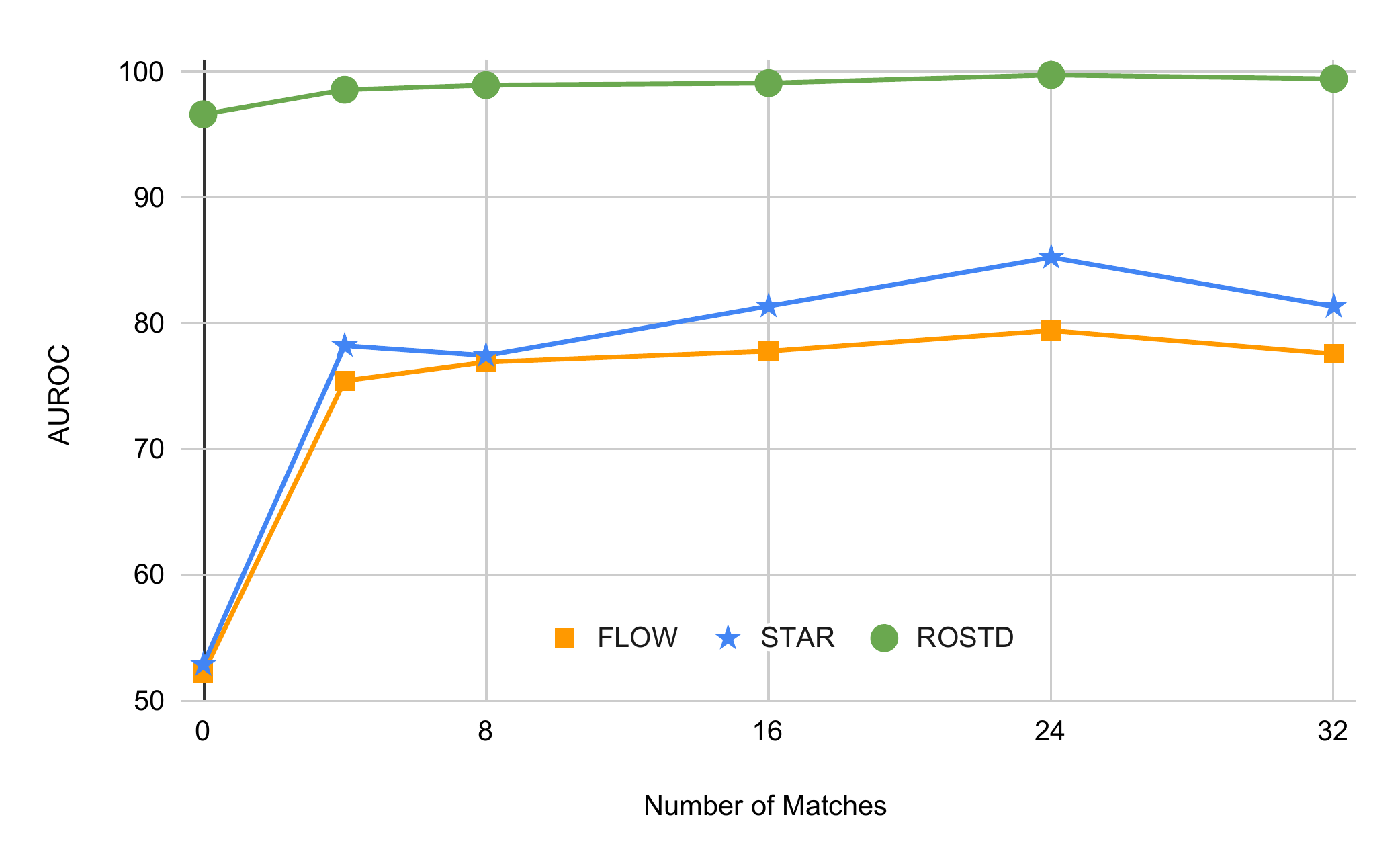}
  \caption{AUROC across datasets as the number of matches increases. A setting of $d=8$ means for each seed example, 8 augmented examples are generated.}
  \label{fig:match_auroc}
\end{figure}

\paragraph{How much does augmentation help?}
Given the extra labels from the seed set, it is natural to ask whether the augmented data add any value. Furthermore, if the augmented data are useful, then we might want to know what an ideal number of additional datapoints would be.
Figure~\ref{fig:match_auroc} displays the AUROC of a model trained on varying the number of augmented datapoints, where ``0'' represents including only known OOS examples.  We see a 
trend that accuracy improves for all target datasets as we add more pseudo-labeled examples, showing that augmentation helps. Improvement reaches a max around 24 matches per seed example, which suggests that the benefit of adding more datapoints has a limit. Accordingly, we use 24 matches for all results listed in Table~\ref{tab:joint}.  

\paragraph{Does the extraction technique matter?}
Previous sections have established that extracting matches based on maximizing similarity to known OOS examples might not be ideal.  We now ask what would happen if we went to the extreme by extracting matches that have no discernible relation to known OOS examples.  The final row of Table \ref{tab:joint} reveals the result of using random selection as an extraction technique.  While Random is not always the worst, its poor performance across all metrics strongly suggests that augmented data should have at least some connection to the original seed set. 

\paragraph{Is filtering even necessary?}
Since the source data distribution is obviously distinct from the target data distribution, perhaps it is possible to bypass elections and simply accept all candidates as OOS, similar to Outlier Exposure from ~\citet{hendrycks2019outlier}. As seen in row 4 of Table~\ref{tab:additional}, we observe that skipping elections leads to a drop in the AUROC of all the models on all datasets.  The effect is most pronounced for STAR, where some of the QQP dialogues overlap with in-scope STAR domains.


\begin{table}
\centering
\resizebox{0.9\linewidth}{!}{
\begin{tabular}{l|cccc}
\Xhline{1pt}
\textbf{Methods} & \textbf{STAR} & \textbf{FLOW} & \textbf{ROSTD} \\
\hline
GloVe w/ QQP          & 0.3885 & 0.3173 & 0.9884 \\
  \qquad \ \ \ w/ US  & 0.2186 & 0.1514 & 0.9633 \\
  \qquad \ \ \ w/ MQA & 0.2722 & 0.2283 & 0.9873 \\
\hline
(4) No Election   & 0.2487 & 0.2548 & 0.9312 \\
(5) Tiny Seed Set & 0.1983 & 0.2111 & 0.8856 \\
(6) Swap Last     & 0.3678 & 0.2980 & 0.9656 \\
    
\hline
\Xhline{1pt}
\end{tabular}}
\caption{Additional AUROC results with data augmented from QQP source and GloVe extraction} \label{tab:additional}
\end{table}

\subsection{Applicability}

\paragraph{How well would a direct classifier perform?}
Indirect prediction is often necessary in real-life because while in-scope data may be trivial to obtain, out-of-scope data is typically lacking.  Accordingly, we artificially limited the amount of data available to mimic this setting.  If such a limitation were to be lifted such that a sufficient amount of known OOS data were available, we could train a model to directly classify such examples.  The first row in Table~\ref{tab:joint} shows the results of using all the available OOS data to perform direct prediction and represents an upper-bound on accuracy.  This also shows there is still substantial room for improvement.

\paragraph{When does \ourmethod~help the most?}
\ourmethod~depends on a small seed set to perform data augmentation, so if this data is unavailable or extremely sparse, then the method will likely suffer. To test this limit, we train a model with half the size of the seed data and double the number of matches ($d=24 \rightarrow 48$) to counterbalance the effect. Despite having an equal amount of pseudo-labeled OOS examples, the model with a tiny seed set (row 5 in Table~\ref{tab:additional}) severely underperforms the original model (row 1). 
Separately, we note that dialogue breakdowns are more likely in conversations that contain multiple turns of context, like in STAR, as opposed dialogues consisting of single lines, as in ROSTD.  Given the more prominent gains by our method in STAR, we conclude that \ourmethod~achieves its gains partially from being able to recognize dialogue breakdowns.

\paragraph{What attributes make a source dataset useful?}
In studying Figure~\ref{fig:source}, we find that the most consistent single source dataset is QQP, which we use as the default for Table~\ref{tab:additional}. 
Reading through some examples in QQP, the pattern we found was that many of the samples contained reasonable, but unanswerable questions that were beyond the skillset of the agent.  One method for curating a useful source dataset then is to look for a corpus containing questions your dialogue model likely cannot answer.  Furthermore, 
PersonaChat (PC) performed particularly well with STAR, a task-oriented dataset. We believe that since goal-oriented chatbots aim to solve specific tasks rather than engage in chit-chat, open-domain chat datasets serve as a good source of OOS examples. 

The themes above suggest that good source datasets are simply those sufficiently different from the target data.  We wondered if there was such as a thing as going to `far', and conversely if there was any harm in being quite `close'.  Concretely, we expected a dataset containing medical questions would represent a substantially different dialogues compared to our target data~\cite{benabacha2019medquad}. Table~\ref{tab:additional} presents results when training with source data from a medical question-answering dataset (MQA) or from unlabeled samples (US) from the same target dataset. The results show a significant drop in performance, indicating that augmentations far away from the decision boundary might not add much value.  Rather, pseudo-labels near the border of INS and OOS instances are the most helpful.  (Further analysis in Appendix~\ref{section:random})


\paragraph{How does one create good OOS examples?}
As a final experiment, we replace only the last utterance with a match when generating candidates, rather than swapping any user utterance. We speculate this creates less diverse pseudo-examples, and therefore decreases the coverage of the OOS space. Indeed, row 6 in Table~\ref{tab:additional} reveals that worse candidates are generated when only the final utterance is allowed to be replaced.
In conjunction with the insight from Section~\ref{subsection:results} that generated examples should be sufficiently different from given OOS examples, we believe that the key to producing good pseudo-OOS examples is to maximize the diversity of fake examples. OOS detection is less about finding out-of-scope cases, but rather an exercise in determining when something is not in-scope.  This subtle distinction implies that the appropriate inductive biases should aim to move away from INS distribution, rather than close to OOS distribution.

%% file: sections/7_conclusion.tex
\section{Conclusion}
This paper presents \ourmethod, a method for improving OOS detection when limited training examples are available by leveraging data augmentation.  Rather than relying on a separate model to support the detection task, our proposed method directly trains a model to detect out-of-scope instances.  Compared to other data augmentation methods, \ourmethod~takes advantage of auxiliary data to expand the coverage of out-of-scope distribution examples rather than trying to extrapolate from in-scope examples.  Moreover, our analysis reveals key techniques for further diversifying the training data to support robustness and prevent overfitting.  

We demonstrate the effectiveness of our technique across three dialogue datasets, where our top models outperform all baselines by a large margin.  Future work could explore detecting more granular levels of errors, as well as more sophisticated methods of filtering candidates~\cite{welleck2019nli}.


%% file: sections/8_appendix.tex
\section{Additional Results}
This section shows the AUPR results corresponding to the AUROC results presented in the main paper.

\begin{figure}[H]
  \includegraphics[width=\linewidth]{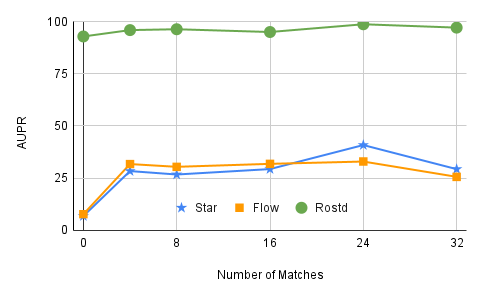}
  \caption{AUPR on with GloVe extraction method as we vary the number of matches. Compare with Figure~\ref{fig:match_auroc} in the main paper.}
  \label{fig:match_aupr}
\end{figure}

We note that the trend is very similar, but just slightly harder to read since the range on the y-axis is larger.  Overall, we reach the same conclusion that augmenting the examples certainly provides a benefit over simply training on the seed data alone.

\section{Latency Impact}
Since \ourmethod~is a data augmentation method, an OOS detector trained with this method incurs no additional cost during inference.  In contrast, Probability Threshold methods will experience extra latency, albeit only minimally, from calculating whether an example falls below the threshold.  Separately, the Outlier Distance methods must measure the distance to multiple clusters which takes a bit of time.  Additionally, the Dropout method must pass the input through $N$ models that form the Bayesian ensemble, leading to much slower inference.

With that said, our OOS detector only performs binary classification.  So if it were to be deployed in a real-world task, such as intent classification, there would need to be an additional downstream model that separately classified the intents when the OOS detector labels a dialogue as in-scope.  To mitigate this issue, a simple solution could be running the intent classifier alongside the OOS detector. Thus, rather than waiting for the result of the detector to start the prediction, the classifier would run in parallel and the classification results would be used only when the detector deemed it necessary.

\section{Source Dataset vs. Technique}
\label{section:random}
One might be curious to know whether choosing a source data or a technique is more important. Before answering this, we first note that source datasets (such as MIX) are not directly comparable to extraction techniques (such as GloVe) since they are different directions to improve performance. Source datasets impose the set of options to choose from, whereas extraction techniques determine how you select the options from that set.  Both decisions can be combined together, and are not mutually exclusive. 

With that said, there is some evidence that choosing the appropriate source dataset can make a more substantial impact.  As initial evidence, notice that the Random extraction technique performs surprisingly well.  This suggests that the gains come largely from using an advantageous source dataset that contains dialogue related examples near the INS and OOS border.  Thus, Random extraction will naturally select some data points near the border as well, and do decently well. In contrast, Section 6.2 compares two new source datasets (MQA and US) that are not near the border, so Random selection of these points should cause the model to do poorly.

\begin{table}
\centering
\resizebox{\linewidth}{!}{
\begin{tabular}{l|ccc}
\Xhline{1pt}
\textbf{Methods} &  \textbf{STAR}  & \textbf{FLOW} & \textbf{ROSTD} \\
\hline
Random w/ MIX (default) & 0.438 & 0.289 & 0.989 \\
\hdashline
Random w/ MQA           & 0.245 & 0.176 & 0.979 \\
Random w/ US            & 0.209 & 0.142 & 0.958 \\  
\Xhline{1pt}
\end{tabular}}
\caption{AUPR results with varying source datasets and Random extraction technique} \label{tab:rand}
\end{table}

To verify this, we ran an additional experiment which extracted MQA samples using a Random approach rather than using GloVe as done originally.  Table~\ref{tab:rand} reveals that indeed AUPR drops noticeably across all datasets.  Similar decreases emerge when the experiment is run on the US dataset as well.  Therefore, we conclude that selection of the source dataset can be fairly critical to success.